# Distance metric learning based on structural neighborhoods for dimensionality reduction and classification performance improvement

Mostafa Razavi Ghods

Department of Computer Engineering, Mashhad Branch, Islamic Azad University, Mashhad, Iran

mrazavi@mshdiau.ac.ir

Mohammad Hossein Moattar (corresponding author)

Department of Computer Engineering, Mashhad Branch, Islamic Azad University, Mashhad, Iran

moattar@mshdiau.ac.ir

Yahya Forghani

Department of Computer Engineering, Mashhad Branch, Islamic Azad University, Mashhad, Iran

yahyafor2000@yahoo.com

**Abstract**

Distance metric learning can be viewed as one of the fundamental interests in pattern recognition and machine learning, which plays a pivotal role on the performance of many learning methods. One of the effective methods in learning such a metric is to learn it from a set of labeled training samples. The issue of data imbalance is the most important challenge of the recent methods. This research tries not only to preserve the local structures but also covers the issue of imbalanced datasets. To do this, the proposed method first tries to extract a low dimensional manifold from the input data. Then, it learns the local neighborhood structures and the relationship of the data points in the ambient space based on the adjacencies of the same data points on the embedded low dimensional manifold. Using the local neighborhood relationships extracted from the manifold space, the proposed method learns the distance metric in a way which minimizes the distance between similar data and maximizes their distance from the dissimilar data points. The evaluations of the proposed method on numerous datasets from the UCI repository of machine learning, and also the KDDCup98 dataset as the most imbalance dataset, justify the supremacy of the proposed approach in comparison with other approaches especially when the imbalance factor is high.

## 1. Introduction

Distance metric learning (DML) for many years has been considered as one the main research interests in works which try to define the similarity and dissimilarity criteria between patterns. Distance metric learning approaches are employed to define an appropriate metric which can reflect the similarity and the dissimilarity of the data points with respect to the application in which they are used. The goal of distance metric learning is to find a real-valued metric function of data pairs under which the data pair with the same label are as close and the data pair from different classes are as far as possible. In this work, the main goal is to learn a function which can transform the input data onto the learned manifold with the least possible amount of changes in the relative distance of data-points from the same class [1].

The application of the distance metric learning the in pattern recognition includes algorithms such as k-means, k-nearest neighbors and kernel-based algorithms such as support vector machines (SVMs) [2]–[9]. Distance metric learning approaches can be categorized into three classes of: fully-supervised, unsupervised, and semi supervised methods. In fully-supervised learning, the ultimate goal is to use the class discriminative information between the data-pairs in order to keep all data within a class as close and the data from different classes as diverse as possible. Zhang et al. [10] have shown that learning the distance metric based on the class discriminative information usually shows better performance than using the classical Euclidean distance.

Supervised distance metric learning itself could be divided into the two categories of local and global approaches. An approach is to learn a global distance metric from the training data in order to satisfy the constraints between all data-pairs simultaneously [5], [11]. The most expressive work in this field is Xing's [11] algorithm which learns a distance metric in the global scale where the distances between the data-pairs are in turn minimized and maximized under the equivalence and inequivalence constraints, respectively. Equivalence and inequivalence constraints may conflict when the data from different classes have multiple distributions. Thus, it is hard to satisfy the whole constrains in the global scale. In order to confront with this phenomenon, local distance metric learning approaches, which take account of the local constraints, are introduced [12]–[14]. These local algorithms only consider the pairwise constraints while avoiding the conflicting ones.

The aforementioned approaches try to present one single metric for all instances of the data. However, learning only single metric may have the deficiencies like: (1) is barely probable to find a metric appropriate for all the training data; (2) a local metric may not be immune to noisy data; (3) a local metric cannot be used in the multi-modal problems. Therefore, it is recommended to use different metrics for multiple distributions of the training data [4], [14], [15].

Generally, supervised distance metric learning could be divided into the two groups of local and global approaches. The local methods could also be subcategorized into the single-metric and multiple-metric approaches. The global methods try to keep the similar samples as close and dissimilar samples as far as possible. Xing's algorithm [11] is a good representative of global approaches which optimizes some equal and inequality constraints simultaneously using the convex optimization methods.

The advantage of using the global approaches is in their ability to capture the distributions from different classes when all the samples belonging to the same class do not obey the same distribution. However, the global approaches may not be able to learn the optimal distance metric when data have multimodal distributions.

Local approaches use the neighborhood information to cope with the multimodal distribution problems. Linear Fisher Discriminant Analysis (LFDA) [13], according to the local information, give more weight to the pairwise neighborhood constraints. Yang et al. [3], proposed a probabilistic approach to optimize the local pairwise constraints. Goldberger et al. [12] utilized a stochastic variant of the KNN classifier to calculate the leave-one-out classification error.

Dimensionality reduction (DR) approaches try to find a low dimensional representation of the data in order to satisfy some goals. Size reduction of the feature vectors for data compression (from the unsupervised perspective) as well as avoiding the curse of dimensionality (from the supervised view) are two main objectives of the dimensionality reduction approaches. However, problems happen when the number of data-points is not sufficient to cover the whole initial high dimensional space. Data visualization is one other goal of the DR approaches, in which the DR is used to project the high dimensional data onto a space with at most two or three dimensions in a way which is comprehensible and visualizable. In data classification application, the DR methods could be used to find a low-dimensional manifold on which the data with the same label are compact, while the data from different classes are discriminant with respect to each other, which itself improves the classification accuracy.

In this article, the proposed method tries to cover a triple of the challenges in the distance metric learning dimensionality reduction. This research is important as it tries to learn the distance metric in such a way that after transformation, which is done by the learned metric function, the data from the same class are as close and the data from different classes are as far from each other as possible. Besides, nowadays many of the real world datasets are found to be imbalanced in terms of the number the points associated to different classes. Thus, the proposed method tries to learn the distance metric with respect to this phenomenon. Furthermore, one other goal of the proposed method is to learn the distance metric in a way that it could be used in any application, independent from the presence or absence of the labeling information.

In this study, we have attempted to learn a low-dimensional manifold out of the data in the initial space. Then, similar, dissimilar and irrelevant data-points are found based their local neighborhood on the manifold. Consequently, based on these neighborhood relationships which are found on the manifold and based the coordinates of the data points on the initial space, distance metric learning is done using a Mahalanobis distance metric learning approach.

The remainder of this paper is organized as follows. Section 0 primarily deals with the concept of distance metric learning and dimensionality reduction followed by some discussion on the different manifold learning approaches. The proposed method will be introduced in section 0. Section 0 describes the experimental setup and analyzes its performance and summarizes its results and finally, section 0 discusses the main findings and concludes this study besides giving some directions for future studies.

## 2. Materials and methods

Distance metric plays a key role in the success of many machine learning algorithms. For example, the classification techniques such as the k-nearest neighbors [16] and the clustering approaches like k-means algorithm are highly dependent on the applied metric in order to model the structural models between the input data. A tangible example in this field could be the visual object recognition problem. Lots of the applications in machine learning could be considered as implicit distance metric learning approaches which are capable of learning the similarities and dissimilarities between visual input objects. In this section we will touch upon some basic ideas about the distance metric learning and dimensionality reduction approaches. Then we will discuss about some of the most promising approaches in this discipline and finally we will conclude this section with a short review on each of the triple of the distance metric learning approaches as supervised, unsupervised and the semi-supervised.

### 2.1. Unsupervised distance metric learning approaches

The unsupervised methods of distance metric learning do not require any supervision data, i.e., they learn the distance metric merely by having the input data coordinates matrix $X$ in such a way that an optimality or discrimination is achieved. In *Equation 1* the unsupervised methods learns from $J(D)$ data by $\lambda_1 = 0$.

$$J(D) = \lambda_1 L(D) + \lambda_2 U(D) \qquad \textit{Equation 1}$$

In which *L(d)* and *U(D)* are labeled and unlabeled data points, respectively. In the following we will talk about some of the most well-known unsupervised approaches of distance metric learning.

**2.1.1. Autoencoder**

Autoencoder is a type of neural network with a generally narrow (bottleneck) hidden layer. This network tries to reconstruct the input data in the output and generally is used for novelty detection and deep learning [17]. This network initially encodes the input and then decodes it to reconstruct it in the output. The goal of the autoencoder as it can be seen in Figure 1 is to reconstruct the input itself.

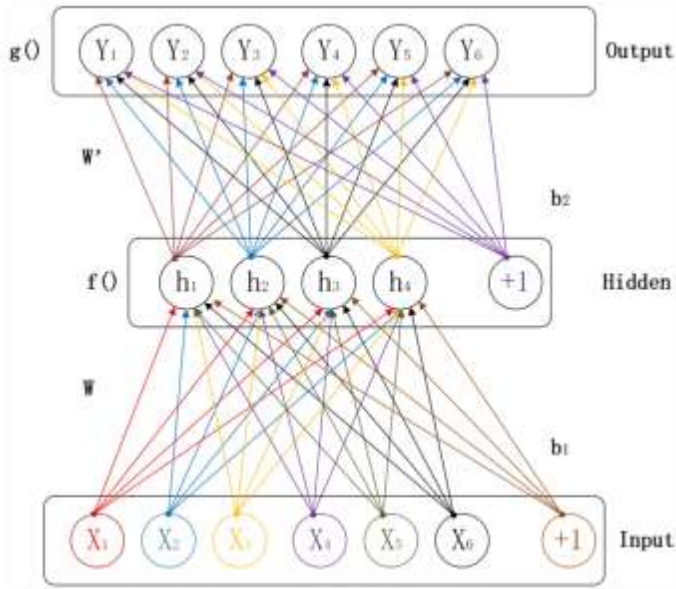

Figure 1. The structure of an autoencoder.

The autoencoder tries to learn the function $S(.)$ as follows:

$$S_{W,W',b_1,b_2}(X) \approx X \qquad\qquad Equation\ 2$$

In which $W, W', b_1, b_2$ are the model parameters. $W$ is a weighted matrix connected to the input and hidden layers and $W'$ is the output layer weights matrix. $b_1$ and $b_2$ are also are the bias vectors of the hidden and output layers, respectively. $S(.)$ is divided into two phases. Phase 1, or the encoding phase is from the input to the hidden layer (*Equation 3*) and the second phase or the decoding phase is from the hidden layer to the output (*Equation 4*). Autoencoder finds the latent space (i.e. hidden variables) embedded in the input data, from the outputs of the hidden layer as denoted by $h$ in Eq. 3.

$$h = f(W \times X + b_1) \qquad \text{Equation 3}$$

$$Y = g(W' \times h + b_2) \qquad \text{Equation 4}$$

In practice, we could use the tied weight $W' = W$ to reconstruct the input $X$ i.e., $Y \approx X$. To do this we could use the square error (*Equation 5*) and cross entropy loss function (*Equation 6*).

$$L_s(W, W', b_1, b_2; X) = \frac{1}{2} Y - X{\wedge}2 \qquad \text{Equation 5}$$

$$L_c(W, W', b_1, b_2; X) = -[X \log Y + (1 - X) \log(1 - Y)] \qquad \text{Equation 6}$$

In these equations, if $X$ is a matrix with real values then we would usually use the Least square loss function and in case the values are binary then the use of the cross entropy loss function would be more appropriate. $Y$ could be calculated from the combination of equations (*Equation 3*) and (*Equation 4*) as follows:

$$Y = g(W' \times X + b_1) + b_2) \qquad \text{Equation 7}$$

Generally, in order to control the weights' scale and to stop the overfitting the regularization term is added to the loss functions *Equation 5* or *Equation 6* where the loss function would be finally as follows.

$$L(W, W', b_1, b_2; X) = L_t(W, W', b_1, b_2; X) + \frac{\lambda}{2} \sum_{l=1}^{nl} \sum_{i=1}^{sl} \sum_{j=1}^{sl+1} \left(W_{ij}^l\right)^2 \qquad \text{Equation 8}$$

In which $L_t$ shows the squared error $L_s$ or the cross-entropy $L_c$. Additionally, $nl$ shows the layer number and $sl$ and $sl + 1$ show the units on the $l$th and $l + 1$th layer, respectively.

### 2.1.2. Locally Linear embedding (LLE)

LLE [18] is another approach to achieve the embedded space which tries to preserve the local neighborhoods of the input data. The difference between LLE and the LE [19] is in the way that they calculate the neighborhoods between the points. LLE is based on the assumption of linear neighborhood between the points, which assumes that each point, $x_i (i = 1, 2, ..., n)$, could be reconstructed using the location of its neighbors, $N_i (i = 1, 2, ..., n)$.

$$\min_{\omega_{ij}} \sum_i \left\| x_i - \sum_x \omega_{ij} x_j \right\|^2 \qquad \text{Equation 9}$$

$$s.t. \sum_j \omega_{ij} = 1 \ (\forall \ i = 1, 2, ..., n)$$

In the second step, LLE tries to retrieve the mappings in a lower dimension while preserving the local relations by solving the optimization problem in *Equation 10*.

$$\min_{\{y_i\}_{i=1}^n} \sum_i \left\| y_i - \sum_x \omega_{ij} y_j \right\|^2 \qquad \text{Equation 10}$$

$$s.t. \sum_{i=1}^n y_i = 0, \ \sum_{i=1}^n y_i y_j = nI$$

LLE is also a local and non-linear method. In this approach, like LE the learned distance between $x_i$ and $x_j$ is the Euclidean distance between $y_i$ and $y_j$. The computation method used in the LLE utilizes both quadratic programming and eigen analysis. Generalization of the LLE for the out-of-sample data is not easy as it calculates the mapped coordinates of the data directly and without calculating any explicit mappings.

### 2.1.3. Isometric feature mapping (Isomap)

Isomap [20] is another approach for learning the low-dimensional spaces where the geodesic distances are devised on a weighted graph with classical scaling (metric Multidimensional Scaling [21]). The main difference between the Isomap, LE and LLE is in their approach of learning the similar data-pairs. In Isomap, in addition to the similarity, the distance between the data-pairs (i.e., the dissimilarities) are first calculated, and then the classic MDS approach is used to calculate the coordinates of the mappings in a way that the pairwise distances are preserved with the best way possible.

Here, the distance between the data-pairs are measured as followed. First, a connected neighborhood graph is constructed on the dataset, this graph could be weighted or unweighted. Then the geodesic distances would be the shortest path between the data-pairs. These computations could be considered as the discrete approximation of the real geodesic distances of the data-pairs on the manifold. Thus, Isomap is a nonlinear and global approach. The learned distance is measured by the Euclidean distance on the low-dimensional space. The computation method used in Isomap is Eigen decomposition. As in Isomap the mapped coordinates of the data are learned directly and without any explicit mappings; thus, like LE and LLE, it is not that easy to extend the Isomap to the out-of-sample data. Geodesic distance has been previously applied successfully for dimensionality reduction in classification and clustering application [22].

### 2.2. Supervised distance metric learning approaches

Supervised distance metric learning algorithms, which preform the learning process based on the data points and their corresponding labels, are discussed in this section. Referring to *Equation 1*, the supervised approaches perform $J(D)$ with $\lambda_2 = 0$. Like the unsupervised approaches, we divide the supervised approaches to different categories based on their characteristics.

### 2.2.1. Linear discriminant analysis (LDA)

LDA [23] is one the popular supervised embedding approaches. This approach, searches for the directions where the data belonging to different classes are discriminated in the best way possible. To be more precise, with the assumption that the data are from C different classes, LDA defines the compactness and separation matrices as follows:

$$\sum_C = \frac{1}{C} \sum_c \frac{1}{n_c} \sum_{x_i \in c} (x_i - \bar{x}_c)(x_i - \bar{x}_c)^T \qquad \text{Equation 11}$$

$$\sum_S = \frac{1}{C} \sum_c (\bar{x}_c - \bar{x})(\bar{x}_c - \bar{x})^T \qquad \text{Equation 12}$$

The goal of LDA is to find $W$ which could be calculated by solving the following equation:

$$\min_{W^T W = I} \frac{tr(W^T \sum_C W)}{tr(W^T \sum_S W)} \qquad \text{Equation 13}$$

By extending the numerator and denominator of *Equation 13*, it could be seen that the numerator corresponds to the sum of the distances between data points and its class center after the mapping, and the denominator corresponds to sum of the distances between the center of each class and the total mean of the data after projection. Therefore, by minimizing *Equation 13* the inter-class scatter increases and at the same time the intra-class scatter decreases. As it is hard to solve *Equation 13*, some researchers [24], [25] have conducted some research on this problem. LDA is a linear global approach. The learned distance between $x_i$ and $x_j$ is the Euclidean distance between $W^T x_i$ and $W^T x_j$. The generalization of LDA to the out-of-sample data is easy as it learns the transformation matrix $W$ explicitly through eigenvalue decomposition.

### 2.2.2. Discriminative Least Squares Regression (DLSR)

discriminative least square approach proposed in [26] is a framework for computing the least square regression (LSR) for multiclass classification. The main goal of this approach is to enlarge the distances between different classes under the framework of the LSR. To do so, [26] has utilized a method called the $\epsilon$-dragging to push the regression objective of different classes back in different directions in a way that the distance between different classes is increased. With the assumption of having $n$ training samples $\{(x_i, y_i)\}_{i=1}^{n}$ in $c (\geq 2)$ classes, where $x_i$ is a datapoint in $\mathbb{R}^m$ and $y_i \in \{1, 2, \ldots, c\}$ is the label of $x_i$. The main goal of the DLSR is to learn the following linear function:

$$y = W^T x + t \qquad \text{Equation 14}$$

Note that an arbitrary set of $c$ independent vectors in $\mathbb{R}^c$ is capable of identifying $c$ classes independently. Thus, 0/1 class label vectors cloud be used as the regression objective for the multiclass classification. In other words, for the $j$th class, $j = 1, 2, \ldots, c$, $f_j = [0, \ldots, 0, 1, 0, \ldots, 0]^T \in \mathbb{R}^c$ could be defined by making the $j$th element equal to one in a way that for $n$ training examples we would have:

$$f_{y_i} \approx W^T x_i + t, i = 1, 2, \ldots, n \qquad \text{Equation 15}$$

Where $W$ is a transformation matrix in $\mathbb{R}^{m \times c}$ and $t$ is a translation vector in $\mathbb{R}^c$. In order to develop a compressed optimization method for multiclass classification, assume that $B \in \mathbb{R}^{n \times c}$ be a constant matrix where the $i$th row and the $j$th column and is defined as follows:

$$B_{ij} = \begin{cases} +1, & \text{if } y_i = j \\ -1, & \text{otherwise.} \end{cases} \qquad \text{Equation 16}$$

From the geometrical viewpoint, each element in $B$ corresponds to a dragging direction. In other words, "+1" indicates the dragging towards the positive direction, whereas "-1" shows the dragging in the negative direction. By performing the mentioned dragging on each element of $Y$ and recording this epsilon with matrix $M$, we would have the following equation:

$$XW + e_n t^T - (Y + B \odot M) \approx 0 \qquad \text{Equation 17}$$

Where $\odot$ indicates the Hadamard (or elementwise) multiplication and is $e_n = [1, 1, \ldots, 1]^T \in \mathbb{R}^n$ a vector of ones.

Now by obtaining the regularized framework of the LSR, we would have the following learning model:

$$\min_{W,t,M} \|XW + e_n t^T - Y - B \odot M\|_F^2 + \lambda \|W\|_F^2 \qquad \text{Equation 18}$$

Where $\lambda$ is a positive regularization term and $\|.\|_F$ indicates the Ferobenius norm.

By adding the term $B \odot M$ in *Equation 18* which is related to the $\epsilon$-dragging for enlarging the inter-class distances, this model could be used for a constrained optimization problem. Based the convex optimization theory, the convexity of *Equation 18* could be easily justified [26] and on this basis it would have one unique answer. For more details on the DLSR algorithm, one could refer to [26]. The $\epsilon$-dragging method is applied as one of the key ideas in our proposed approach.

## 3. The proposed method

This section will describe the proposed method of distance metric learning in detail. The proposed method tries to learn the distance metric in way that the structures between the data-points are preserved as much as possible. In this approach, in order to encounter with the problem of the imbalanced distributions of different classes, for each given data point, two neighborhoods are created, each of which consisting of the data with similar and dissimilar labels to the given data point, respectively. The proposed method tries to preserve the spatial locality of the similar data in relation with each other and to push back the dissimilar data from each data-point. On this basis, and with respect to the fact that the number of the data points in the similar neighborhood is equal to the number of points in the dissimilar neighborhood, the problem of the imbalanced data distributions could also be covered.

As it can be seen in

Figure 2, in the proposed method, in order to increase the manifold and distance metric learning speed besides reaching a feasible amount of system memory on today's computers, first the number of the training data is down-sampled, otherwise the size of the similarity matrix would as big and bulky that it could not be implemented and, as a result, the execution of the proposed method would not be possible. In order to encounter with such big data, a uniform random sampling of the training data is preformed through which the share of each class in training samples will be remained intact. The down-sampling factor is considered to be 0.1 of all samples.

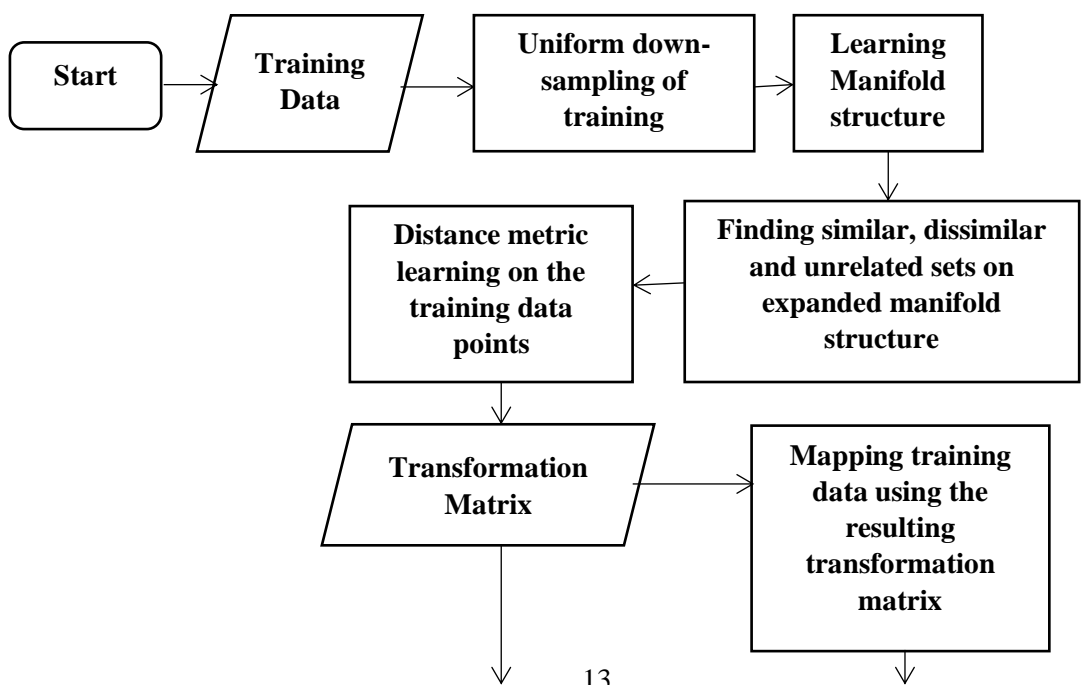

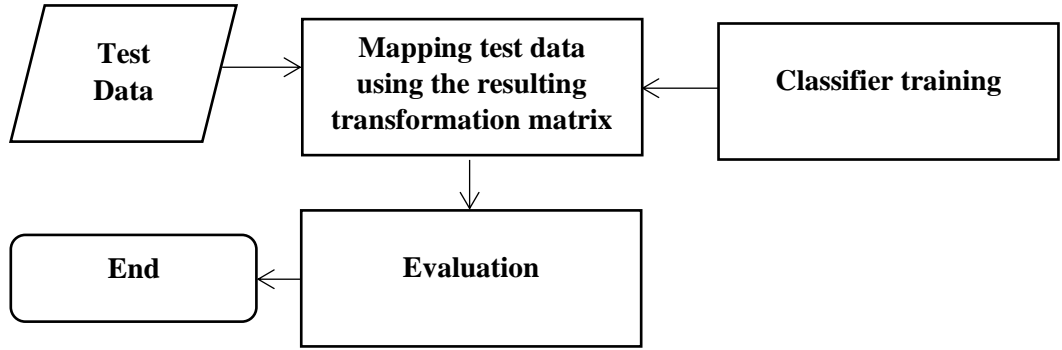

Figure 2. Overall process of the proposed method.

After sample reduction, manifold learning is conducted on these data using one of the manifold learning approaches in order to extract the local neighborhoods of the nodes based on their adjacencies on the manifold. Consequently, based on these extracted local neighborhoods, two neighborhoods are created for each given data point. As it can be seen in Figure 3, one of the created neighborhoods is dedicated to the data with the same label whereas the other neighborhood consists of the dissimilar neighbors to the given data point. Other data points are regarded as so called unrelated set. Finally, as it is depicted in Figure 4, distance metric learning based on the initial coordinates of the given data point in ambient space and with respect to the similarity and dissimilarity relations thanks to constructed similar and dissimilar neighborhoods is conducted in a way that the similar data points to the given point would be more close to it than the other dissimilar points.

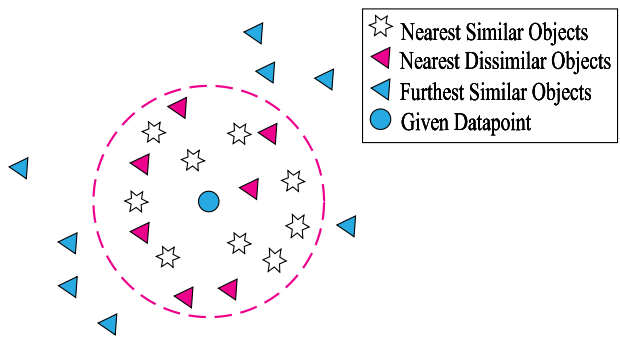

Figure 3. The local patch consisting of the dissimilar neighbors.

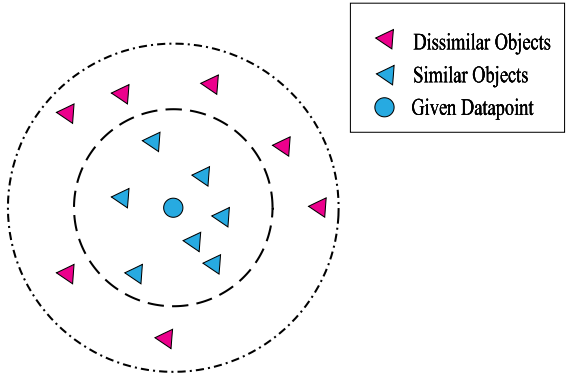

figure 4. The local neighborhoods after distance metric learning with the proposed approach.

After discriminating the similar and dissimilar neighborhoods, as well as the unrelated data points which are not contained in either of similar and dissimilar neighborhoods, they are ordered as shown in Figure 5, based on their distance to the given data point and the following relation vector is created.

| $x_i$ | $S_i$ | $D_i$ | $U_i$ |

Figure 5. The representation of data points after manifold embedding and similarity calculation.

In Figure 5, $x_i$ shows the given data point, $S_i$ shows the faraway points from $x_i$ with the similar class labels and, $D_i$ shows the neighbors with the dissimilar data and $U_i$ indicates the unrelated data which are not included in either of the similar and dissimilar sets with respect to the given data point. In the other words, if a data point is not a member of either of their similar or dissimilar neighborhoods, it is said to be unrelated.

At this stage one of the distance metric learning methods e.g., the Mahalanobis distance, could be used. In the proposed framework, we have adopted the Discrete Least Square Regression (DLSR), proposed in [26] and modified the approach in it in order to be compatible with the proposed distance metric learning. Having the above similar/dissimilar/unrelated sets the proposed approach can be formulated as the following optimization problem as inspired from [26]:

$$\min\|XW + e_n t^T - Y - B \odot M\|_F^2 + \lambda\|W\|_F^2 \qquad \text{Equation 19}$$

In our proposed method, $X \in \mathbb{R}^{n \times m}$ is the input data matrix, $W \in \mathbb{R}^{m \times n}$ is the transformation matrix to the similarity space (resulted from the distance metric learning) and $e_n = [1, 1, \ldots, 1]^T \in \mathbb{R}^n$ is a vector consisting of ones. Also, $Y \in R^{n \times n}$ and $B \in R^{n \times n}$ are two constant matrices each of which are in the $i$th row and the $j$th column as follows:

$$Y_{i,j} = \begin{cases} 1, & \text{if } l_i = j, j \in P_{s_i} \\ 0, & \text{otherwise} \end{cases} \qquad \text{Equation 20}$$

$$B_{i,j} = \begin{cases} +1, & \text{if } l_i = j, j \in P_{s_i} \\ -1, & \text{if } l_i = j, j \in P_{d_i} \\ 0, & \text{otherwise} \end{cases} \qquad \text{Equation 21}$$

Where $P_{s_i}$ and $P_{d_i}$ show the similar and dissimilar sets for each given data point. In other words, each element $Y_{i,j}$ will be equal to one in case that the $j$th data point which has the same label as i be in the furthest neighborhood of the given it. Also, matrix $B$ shows the similar/dissimilar/unrelated set (+1,-1 and 0 respectively) information as gathered from the previous stage. The other matrices and variables included in [26], as well as the calculations of transformation matrix, W, are done precisely based on the assumptions contained in the DLSR algorithm [26].

As you can see in

Figure 2, after calculating the mapping matrix W, all the training/test data are mapped to the similarity space using the following equation.

$$X' = X \times W + e_n \times t^T \hspace{4em} \textit{Equation 22}$$

Where $X'$ is the transformed data matrix, showing the data mapped onto the similarity space and also $t \in \mathbb{R}^n$ is a translation vector.

### 3.1. The objective of distance metric learning

As it is shown in Figure 3 and Figure 4, the main objective of a distance metric learning algorithm, is to learn the parameters of the metric which are best suited for the constraints in such a way that it is the best approximation of the distance embedded between the data points. Distance metric learning is commonly expressed as an optimization problem, as the general form below:

$$\min_{M} l(M, S, D, R) + \lambda R(M) \hspace{4em} \textit{Equation 23}$$

Where $l(M, S, D, R)$ is a loss function that acquires a penalty in case the training constraints are violated and $R(M)$ regularizes the parameters $M$ of the learned metric and $\lambda \geq 0$ is a regularization parameter.

After the learning phase, the resulted function is used to improve the performance of a metric-based algorithm, which is most commonly k-Nearest Neighbors (k-NN). The main goal of using the k-NN is to preserve the symmetry in the distance metric learning phase, with the sense that, as seen in Figure 3 and Figure 4, the number of the similar neighbors is equal with the number of the neighbors from other classes around each data point. As a result, the supremacy of the proposed method is that it learns the distance metric in a balanced way as it uses an equal number of the similar and dissimilar data points to learn the distance metric.

## 4. Experiments

This section will make a comparison between the proposed method and the Discriminative Least Squares Regression (DLSR) [26] and some other fundamental methods of dimensionality reduction.

### 4.1. Dataset

In order to evaluate the proposed method in this research the following numeric datasets which are obtained from the UCI repository of machine learning are employed.

Table 1. The properties of the datasets.

| Dataset | #Samples | # Class | # Features | Imbalance ratio |
| --- | --- | --- | --- | --- |
| Vehicle | 846 | 4 | 18 | 1.09 |
| Bupa | 345 | 2 | 6 | 1.37 |
| Glass | 214 | 6 | 8 | 8.44 |
| Ionosphere | 351 | 2 | 34 | 1.78 |
| Iris | 150 | 3 | 4 | 1 |
| KDD | 494021 | 5 | 41 | 7528.03 |
| Monks | 124 | 2 | 6 | 1 |
| New-thyroid | 215 | 3 | 5 | 5 |
| Pima | 768 | 2 | 8 | 1.86 |
| WDBC | 569 | 2 | 30 | 1.68 |
| Wholesale | 440 | 2 | 7 | 2.09 |
| Wine | 178 | 3 | 13 | 1.47 |

In which the imbalance ratio is the proportion of the population of the majority class to the population of the minority class which could be calculated from the following equation.

$$R_{Im} = \frac{n_{major}}{n_{minor}} \qquad \text{Equation 24}$$

Where $R_{Im}$ is the imbalance ratio and $n_{major}$ and $n_{minor}$ are the population of the majority class to the minority class, respectively.

## 4.2. Evaluation criteria

In order to compare the proposed method with other approaches we have employed the following evaluation criteria:

Accuracy or the correct rate is the proportion of the correctly classified data to the total number of the items in the dataset.

$$ACC = \frac{TP + TN}{TP + TN + FP + FN} \qquad \text{Equation 25}$$

Sensitivity, true positive rate (TPR), recall, or the hit rate, is the proportion of the data which are correctly classified in the positive class to the total of the positive data.

$$SEN = \frac{TP}{TP + FN} \qquad \text{Equation 26}$$

Specificity or true negative rate is the proportion of the negative points which are correctly classified in the negative class to the total number of the negative samples.

$$SPC = \frac{TN}{TN + FP} \qquad \text{Equation 27}$$

## 4.3. The evaluation scenarios and experimental results

In this section we will analyze and make a comparison between the performance results of the proposed method and some other well-known approaches of distance metric learning and dimensionality reduction and also the original DLSR algorithm with respect to the evaluation measures. To do this, the 10-fold cross validation is utilized. The results are based on the performance of the two k-NN classifier and SVM classifier with the RBF kernel. The accuracy of different approaches including DLSR and the proposed approach are depicted in Table 2. In these experiments the proposed method has employed different manifold learning approaches such as PCA, LDA, MDS, Isomap, LLE, Kernel PCA and Autoencoder. The experiments are performed for different latent dimensions and the best results are reported in the tables. Note that, in the following tables $(d, R)$ respectively show the best latent dimension and the rank of the method on the corresponding dataset.

As it can be seen in Table 2, from the total of 12 experiments on different datasets, the proposed method of distance metric learning using the LLE, Kernel PCA and LDA approaches for manifold learning has gained the first rank on 7, 6 and 5 datasets, respectively. While, under the same circumstances the other methods such as the pure manifold learning, feature selection and the DLSR have achieved the best accuracy only in one experiment which is still equal to the result of the proposed method.

Therefore, from total of 12 experiments, the proposed framework, has totally gained the first rank, whereas the base approaches have the first rank only in one experiment which is a testimony of the absolute excellence of the proposed approach from the accuracy viewpoint using 7-NN classifier.

Also, with respect to the fact that among different manifold learning methods combined with DML, LLE has gained the maximum rank, it could be concluded that this approach has got the best performance in finding the structural neighborhoods in comparison with the other manifold learning approaches in terms of the accuracy using the 7-NN classifier.

Table 2. Accuracy comparison between different approaches versus the proposed using 10-fold cross validation and 7-NN classifier with (d,r) indicating the best latent dimensionality and the rank of the approach, respectively (AE denotes auto-encoder approach).

| Dataset | Dimensionality reduction | | | | Feature selection | | | The proposed method | | | | | | |
|---|---|---|---|---|---|---|---|---|---|---|---|---|---|---|
| | PCA | LLE | Kernel PCA | AE | Fisher | Gini | DLSR | PCA | LDA | MDS | Isomap | LLE | Kernel PCA | AE |
| Vehicle | 0.6823 (13, 9) | 0.6117 (17, 12) | 0.2588 (9, 14) | 0.5294 (5, 13) | 0.6823 (17, 9) | 0.6705 (17, 11) | 0.9183 (17, 1) | 0.8235 (1, 4) | 0.8705 (1, 3) | 0.8235 (1, 4) | 0.8235 (1, 4) | 0.8941 (13, 2) | 0.8 (1, 8) | 0.8235 (13, 4) |
| Bupa | 0.5714 (1, 12) | 0.6571 (3, 9) | 0.4285 (5, 14) | 0.5714 (3, 12) | 0.6857 (5, 7) | 0.7428 (1, 9) | 0.7573 (3, 2) | 0.6285 (1, 10) | 0.7714 (5, 1) | 0.6285 (1, 10) | 0.6857 (1, 7) | 0.7142 (3, 5) | 0.7142 (1, 5) | 0.7428 (5, 3) |
| Glass | 0.5454 (1, 10) | 0.5454 (5, 10) | 0.5 (7, 12) | 0.4545 (1, 13) | 0.7272 (9, 3) | 0.7272 (9, 3) | NA | 0.7272 (1, 3) | 0.7272 (3, 3) | 0.7272 (1, 3) | 0.7272 (5, 3) | 0.7727 (3, 1) | 0.7727 (5, 1) | 0.7272 (9, 3) |
| Ionosphere | 0.8888 (8, 11) | 0.8055 (22, 14) | 0.9166 (8, 5) | 0.8888 (15, 11) | 0.9166 (15, 5) | 0.9166 (15, 5) | 0.8694 (29, 13) | 0.9166 (15, 5) | 0.9722 (1, 1) | 0.9166 (15, 5) | 0.9444 (15, 4) | 0.9722 (29, 1) | 0.9722 (1, 1) | 0.9166 (29, 5) |
| Iris | 1 (1, 1) | 1(2, 1) | 1(2, 1) | 1 (1, 1) | 1(2, 1) | 1(2, 1) | NA | 1(1, 1) | 1(1, 1) | 1(1, 1) | 1(1, 1) | 1(1, 1) | 1(1, 1) | 1(1, 1) |
| KDD | 0.9879 (1, 10) | 0.9839 (28, 12) | 0.7915 (10, 14) | 0.9819 (37, 13) | 0.9919 (10, 6) | 0.9919 (19, 6) | 0.9901 (28, 9) | 0.9939 (10, 2) | 0.9939 (10, 10) | 0.9939 (1, 2) | 0.9939 (1, 2) | 0.9959 (19, 1) | 0.9939 (10, 2) | 0.9919 (1, 6) |
| Monks | 0.8333 (5, 8) | 0.9166 (5, 4) | 0.8333 (3, 8) | 0.5 (1, 14) | 0.5833 (3, 12) | 0.5833 (3, 12) | 0.7916 (5, 11) | 1(5, 1) | 0.8333 (3, 8) | 1(5, 1) | 0.9166 (5, 4) | 0.9166 (3, 4) | 0.9166 (3, 4) | 1(3, 1) |
| New-thyroid | 0.9545 (5, 1) | 0.9090 (3, 3) | 0.5909 (1, 13) | 0.9090 (3, 3) | 0.9090 (1, 3) | 0.9090 (1, 3) | NA | 0.9090 (1, 3) | 0.9090 (1, 3) | 0.9090 (1, 3) | 0.9090 (1, 3) | 0.9090 (1, 3) | 0.9545 (5, 1) | 0.9090 (1, 3) |
| Pima | 0.7532 (3, 10) | 0.7142 (7, 12) | 0.6493 (3, 14) | 0.6883 (7, 13) | 0.7922 (1, 2) | 0.7922 (1, 2) | 0.7597 (1, 9) | 0.7792 (5, 4) | 0.8051 (1, 1) | 0.7792 (5, 4) | 0.7792 (7, 4) | 0.7792 (1, 4) | 0.7792 (3, 4) | 0.7532 (7, 10) |
| WDBC | 0.9298 (7, 9) | 0.9122 (8, 12) | 0.6315 (22, 14) | 0.8596 (8, 13) | 0.9298 (22, 9) | 0.9298 (22, 9) | 0.9807 (1, 4) | 0.9473 (1, 5) | 0.9473 (1, 5) | 0.9473 (1, 5) | 0.9473 (1, 5) | 0.9824 (15, 1) | 0.9824 (1, 1) | 0.9824 (1, 1) |
| Wine | 0.7777 (4, 9) | 0.9444 (13, 8) | 0.3888 (1, 13) | 0.7222 (1, 12) | 0.7777 (13, 9) | 0.7777 (13, 9) | NA | 1(1, 1) | 1(1, 1) | 1(1, 1) | 1(1, 1) | 1(1, 1) | 1(4, 1) | 1(1, 1) |
| Wholesale | 1(3, 1) | 0.9545 (3, 11) | 0.6818 (3, 13) | 0.7045 (5, 12) | 1(5, 1) | 1(7, 1) | NA | 1(1, 1) | 0.9772 (1, 8) | 1(1, 1) | 1(3, 1) | 1(7, 1) | 0.9772 (1, 8) | 0.9772 (1, 8) |
| Average rank | 7.66 | 9 | 11.25 | 10.83 | 5.67 | 5.5 | 7.14 | 3.33 | 3.08 | 3.33 | 3.25 | 2.08 | 3.08 | 3.92 |

Table 2 denotes the comparison between the proposed methods and other approaches of distance metric learning and dimensionally reduction in term of sensitivity. As it can be seen in Table 3, from the total of 12 experiments on different datasets, the proposed method of distance metric learning using LLE, Auto-encoder and the PCA approaches of manifold learning has gained the first rank on 10, 10 and 9 datasets, respectively. Whereas, under the same circumstances from the other methods, approaches such Auto-encoder, Gini and Fisher has gained the first rank in 7, 6 and 6 experiments, respectively.

Table 3. Sensitivity comparison between different approaches versus the proposed using 10-fold cross validation and 7-NN classifier with (d,r) indicating the best dimensionality and the rank of the approach, respectively (AE denotes auto-encoder approach).

| dataset | Dimensionality reduction | | | | Feature selection | | DLSR | The proposed method | | | | | | |
|---|---|---|---|---|---|---|---|---|---|---|---|---|---|---|
| | PCA | LLE | Kernel PCA | AE | Fisher | Gini | | PCA | LDA | MDS | Isomap | LLE | Kernel PCA | AE |
| Vehicle | 0.95 (13, 9) | 0.75 (4, 14) | 1(1, 1) | 0.95 (9, 9) | 0.9846 (5, 7) | 0.9692 (17, 8) | 1(13, 1) | 0.95 (1, 9) | 1 (1, 1) | 0.95 (1, 9) | 0.95 (5, 9) | 1(1, 1) | 1(1, 1) | 1 (13, 1) |
| Bupa | 0.4667 (3, 13) | 0.6666 (3, 6) | 1 (1, 1) | 0.5333 (5, 9) | 0.75 (5, 4) | 0.8 (1, 2) | 0.7616 (3,3) | 0.5333 (1, 9) | 0.7333 (5, 5) | 0.5333 (1, 9) | 0.5333 (1, 9) | 0.6 (3, 7) | 0.4667 (1, 13) | 0.6 (3, 7) |
| Glass | 0.8571 (1, 8) | 0.8571 (1, 8) | 0.5 (7, 13) | 0.7142 (1, 12) | 0.8571 (5, 8) | 0.8571 (5, 8) | NA | 1(3, 1) | 1(5, 1) | 1(3, 1) | 1(1, 1) | 1(3, 1) | 1(3, 1) | 1(1, 1) |
| Ionosphere | 0.9565 (8, 10) | 0.9565 (22, 10) | 0.9130 (1, 14) | 0.9522 (15, 10) | 1(1, 1) | 1(1, 1) | 0.9913 (29, 9) | 1 (15, 1) | 1(1, 1) | 1 (15, 1) | 1(1, 1) | 1(1, 1) | 0.9522 (1, 10) | 1 (29, 1) |
| Iris | 1(1, 1) | 1(1, 1) | 1(1, 1) | 1(1, 1) | 1(1, 1) | 1(1, 1) | NA | 1(1, 1) | 1(1, 1) | 1(1, 1) | 1(1, 1) | 1(1, 1) | 1(1, 1) | 1(1, 1) |
| KDD | 1(10, 1) | 1(10, 1) | 1 (10, 1) | 1(19, 1) | 0.9919 (10, 14) | 0.9974 (10, 12) | 0.9971 (19, 13) | 1 (10, 1) | 1 (1, 1) | 1 (1, 1) | 1(1, 1) | 1(1, 1) | 1 (10, 1) | 1(1, 1) |
| Monks | 1(5, 1) | 1(1, 1) | 0.8333 (3, 12) | 1(3, 1) | 1(5, 1) | 1(1, 1) | 0.7333 (5, 13) | 1(5, 1) | 0.6666 (1, 14) | 1(5, 1) | 1(5, 1) | 1(3, 1) | 1(3, 1) | 1(3, 1) |
| New-thyroid | 1(5, 1) | 1(1, 1) | 0.8666 (1, 13) | 1(5, 1) | 1(1, 1) | 1(1, 1) | NA | 1 (1, 1) | 1 (1, 1) | 1 (1, 1) | 1 (1, 1) | 1 (1, 1) | 1 (1, 1) | 1 (1, 1) |
| Pima | 0.4444 (3, 13) | 0.4444 (3, 13) | 1(1, 1) | 0.9259 (3, 3) | 0.7037 (1, 3) | 0.7037 (1, 3) | 0.6185 (1, 5) | 0.5185 (1, 8) | 0.5555 (7, 6) | 0.5185 (5, 8) | 0.5185 (1, 8) | 0.5555 (1, 8) | 0.5185 (1, 8) | 0.5185 (5, 8) |
| WDBC | 1(8, 1) | 1(8, 1) | 1 (22, 1) | 1(1, 1) | 1(22, 1) | 1(22, 1) | 0.9861 (1, 1) | 1(1, 1) | 1(1, 1) | 1(1, 1) | 1(1, 1) | 1(1, 1) | 1(1, 1) | 1(1, 1) |
| Wine | 0.85 (1, 10) | 0.8333 (4, 11) | 1(4, 1) | 1(4, 1) | 0.8333 (4, 11) | 0.8333 (1, 11) | NA | 1(1, 1) | 1(1, 1) | 1(1, 1) | 1(1, 1) | 1(1, 1) | 1(1, 1) | 1(1, 1) |
| Wholesale | 1(3, 1) | 0.9666 (3, 11) | 1(3, 1) | 1(1, 1) | 1(5, 1) | 1(7, 1) | NA | 1(1, 1) | 0.9666 (1, 11) | 1(1, 1) | 1(3, 1) | 1(7, 1) | 0.9666 (1, 11) | 1(5, 1) |
| Average rank | 5.75 | 6.5 | 5 | 4.08 | 4.42 | 4.42 | 8.29 | 2.92 | 3.67 | 2.92 | 2.92 | 1.92 | 4 | 2.08 |

In a comparison between the proposed method and the base methods in terms of the specificity according to **Error! Reference source not found.**, from the total of the 12 experiments, thanks to the data mapping on the manifold learnt by the kernel PCA, LDA, and Auto-encoder methods, the proposed method has gained the first rank in 7, 6, and 7 experiments, respectively. Yet, under the same circumstances out of the base methods of manifold learning, the methods of Kernel PCA, Gini index, and Fisher have earned the first rank in 5, 3, and 4 experiments, respectively.

Table 4. Specificity comparison between different approaches versus the proposed using 10-fold cross validation and 7-NN classifier with (d,r) indicating the best dimensionality and the rank of the approach, respectively (AE denotes auto-encoder approach).

| Dataset | Dimensionality reduction | | | | Feature selection | | DLSR | The proposed method | | | | | | |
|---|---|---|---|---|---|---|---|---|---|---|---|---|---|---|
| | PCA | LLE | Kernel PCA | AE | Fisher | Gini | | PCA | LDA | MDS | Isomap | LLE | Kernel PCA | AE |
| Vehicle | 0.9692 (13, 11) | 0.8923 (17, 13) | 1 (5, 1) | 0.8307 (1, 14) | 0.9846 (5, 10) | 0.9692 (17, 11) | 1 (17, 1) | 1(1, 1) | 1 (1, 1) | 1(1, 1) | 1(1, 1) | 1(1, 1) | 1(1, 1) | 1(1, 1) |
| Bupa | 0.7 (1, 11) | 0.65 (3, 13) | 0 (1, 14) | 0.7 (1, 11) | 0.75 (5, 8) | 0.8 (1, 3) | 0.7564 (3, 7) | 0.75 (3, 8) | 0.8 (1, 3) | 0.75 (3, 8) | 0.8(1, 3) | 0.8 (3, 3) | 0.9(1, 1) | 0.9 (5, 1) |
| Glass | 0.6666 (7, 13) | 0.7333 (3, 10) | 0.8 (1, 1) | 0.8 (5, 1) | 0.7333 (1, 10) | 0.7333 (1, 10) | NA | 0.8 (1, 1) | 0.8 (3, 1) | 0.8(1, 1) | 0.8(3, 1) | 0.8 (1, 1) | 0.8(5, 1) | 0.8 (1, 1) |
| Ionosphere | 0.7692 (8, 8) | 0.6153 (15, 14) | 1(8, 1) | 0.7692 (15, 8) | 0.8461 (15, 4) | 0.8461 (15, 4) | 0.7076 (8, 13) | 0.7692 (1, 8) | 0.9230 (1, 2) | 0.7692 (1, 8) | 0.8461 (15, 4) | 0.9230 (29, 2) | 0.8461 (1, 4) | 0.7692 (1, 8) |
| Iris | 1(1, 1) | 1(1, 1) | 1(1, 1) | 1(1, 1) | 1(1, 1) | 1(1, 1) | NA | 1(1, 1) | 1(1, 1) | 1(1, 1) | 1(1, 1) | 1(1, 1) | 1(1, 1) | 1(1, 1) |
| KDD | 0.9906 (10, 4) | 0.9439 (10, 13) | 0.0373 (10, 14) | 1 (37, 1) | 1 (10, 1) | 0.9906 (10, 4) | 0.9953 (37, 3) | 0.9906 (10, 4) | 0.9906 (1, 4) | 0.9906 (1, 4) | 0.9906 (1, 4) | 0.9906 (1, 4) | 0.9906 (1, 4) | 0.9906 (1, 4) |
| Monks | 0.8333 (3, 9) | 0(1, 14) | 0.8333 (1, 9) | 1(3, 1) | 0.6666 (5, 12) | 0.6666 (5, 12) | 0.85 (5, 8) | 1(1, 1) | 1(1, 1) | 1(3, 1) | 1(1, 1) | 0.8333 (1, 9) | 1(1, 1) | 1(1, 1) |
| New-thyroid | 0.8571 (3, 2) | 0.7142 (3, 7) | 0.8666 (1, 1) | 0.8571 (5, 2) | 0.8571 (3, 2) | 0.8571 (3, 2) | NA | 0.7142 (1, 7) | 0.7142 (1, 7) | 0.7142 (1, 7) | 0.7142 (1, 7) | 0.7142 (1, 7) | 0.8571 (5, 2) | 0.7142 (1, 7) |
| Pima | 0.92 (5, 5) | 0.9 (7, 12) | 1 (3, 11) | 0.82 (7, 14) | 0.92 (3, 5) | 0.92 (3, 5) | 0.836 (7, 13) | 0.92 (5, 5) | 0.96 (1, 2) | 0.92 (5, 5) | 0.92 (7, 5) | 0.94 (7, 3) | 0.92 (3, 5) | 0.94 (7, 3) |
| WDBC | 1(1, 1) | 0.7619 (1, 13) | 1(1) | 0.7619 (8, 13) | 0.8095 (8, 11) | 0.8095 (8, 11) | 0.9714 (1, 3) | 0.8571 (1, 7) | 0.8571 (1, 7) | 0.8571 (1, 7) | 0.8571 (1, 7) | 0.9523 (15, 4) | 0.9523 (15, 4) | 0.9523 (1, 4) |
| Wine | 1(4, 1) | 1(1, 1) | 1(1, 1) | 1(1, 1) | 1(1, 1) | 1(1, 1) | NA | 1(1, 1) | 1(1, 1) | 1(1, 1) | 1(1, 1) | 1(1, 1) | 1(1, 1) | 1(1, 1) |
| Wholesale | 1(3, 1) | 0.9285 (3, 12) | 1(1, 1) | 0.0714 (5, 13) | 1(5, 1) | 1(5, 1) | NA | 1(1, 1) | 1(1, 1) | 1(1, 1) | 1(1, 1) | 1(1, 1) | 1(1, 1) | 1(1, 1) |
| Average rank | 5.58 | 10.25 | 3.83 | 6.67 | 5.5 | 5.42 | 6.86 | 3.75 | 2.58 | 3.75 | 3 | 3.01 | 2.17 | 2.75 |

Also, the proposed methods have gained a lower on average rank than the base approaches. Also, we can claim kernel PCA as the best approach in finding the neighborhoods on the manifold in terms of the specificity by using the 7-NN classifier.

Comparing the proposed method with the base approaches in terms of the accuracy using the SVM classifier according to **Error! Reference source not found.**, out of the total of 12 experiments, the proposed method using the LLE, Auto-encoder and the PCA, has gained the first rank in 6, 6, and 5 experiments respectively. Whereas, under the same circumstances, from the base approaches the Fisher, Gini, and PCA have received the first rank in 2, 2, and 2 experiments, respectively.

Table 5. Accuracy comparison between different approaches versus the proposed using 10-fold cross validation and SVM classifier with (d,r) indicating the best latent dimensionality and the rank of the approach, respectively (AE denotes auto-encoder approach).

| Dataset | Dimensionality reduction | | | | Feature selection | | The proposed method | | | | | | |
|---|---|---|---|---|---|---|---|---|---|---|---|---|---|
| | PCA | LLE | Kernel PCA | AE | Fisher | Gini | PCA | LDA | MDS | Isomap | LLE | Kernel PCA | AE |
| Vehicle | 0.4588 (1, 9) | 0.4470 (17, 10) | 0.2588 (1, 13) | 0.3411 (13, 11) | 0.4941 (5, 8) | 0.3058 (5, 12) | 0.8(5, 6) | 0.8235 (1, 2) | 0.8(5, 6) | 0.8117 (5, 4) | 0.8470 (1, 1) | 0.8117 (9, 4) | 0.8235 (13, 2) |
| Bupa | 0.6 (3, 9) | 0.5714 (1, 11) | 0.5714 (1, 11) | 0.5714 (1, 11) | 0.6 (3, 9) | 0.7714 (1, 1) | 0.7142 (1, 2) | 0.6857 (5, 5) | 0.7142 (1, 2) | 0.7142 (1, 2) | 0.6857 (1, 5) | 0.6571 (5, 8) | 0.6857 (3, 5) |
| Glass | 0.5909 (7, 1) | 0.5454 (9, 2) | 0.5454 (9, 2) | 0.3181 (9, 13) | 0.5454 (7, 2) | 0.4545 (5, 12) | 0.5238 (1, 5) | 0.5238 (1, 5) | 0.5238 (1, 5) | 0.5238 (1, 5) | 0.5238 (3, 5) | 0.5238 (3, 5) | 0.5238 (1, 5) |
| Ionosphere | 0.9444 (15, 7) | 0.7222 (22, 12) | 0.9166 (22, 10) | 0.6388 (8, 13) | 0.9722 (8, 1) | 0.9444 (8, 7) | 0.9722 (15, 1) | 0.9722 (1, 1) | 0.9722 (15, 1) | 0.9166 (8, 10) | 0.9722 (15, 1) | 0.9444 (1, 7) | 0.9722 (22, 1) |
| Iris | 1(1, 1) | 1(2, 1) | 0.9333 (2, 11) | 0.6666 (1, 13) | 1(1, 1) | 1(1, 1) | 1(1, 1) | 0.9333 (1, 11) | 1(1, 1) | 1(1, 1) | 1(1, 1) | 1(1, 1) | 1(1, 1) |
| KDD | 0.9779 (1, 9) | 0.9839 (19, 8) | 0.7855 (1, 13) | 0.9338 (37, 12) | 0.9378 (19, 11) | 0.9679 (1, 10) | 0.9939 (1, 3) | 0.9939 (10, 3) | 0.9939 (1, 3) | 0.9919 (1, 7) | 0.9959 (1, 1) | 0.9939 (10, 3) | 0.9959 (1, 1) |
| Monks | 0.8333 (3, 5) | 0.5 (1, 13) | 0.8333 (5, 5) | 0.75 (1, 11) | 0.9166 (5, 2) | 0.9166 (5, 2) | 0.8333 (3, 5) | 0.75 (1, 11) | 0.8333 (3, 5) | 0.9166 (5, 2) | 0.8333 (3, 5) | 0.8333 (3, 5) | 1(3, 1) |
| New-thyroid | 0.7272 (1, 12) | 0.8636 (5, 9) | 0.6818 (1, 13) | 0.8181 (5, 11) | 0.9090 (1, 6) | 0.9090 (1, 6) | 0.9545 (5, 1) | 0.8636 (1, 9) | 0.9545 (5, 1) | 0.9545 (5, 1) | 0.9545 (5, 1) | 0.9545 (5, 1) | 0.9090 (3, 6) |
| Pima | 0.6883 (1, 10) | 0.6623 (7, 11) | 0.6493 (1, 12) | 0.6493 (1, 12) | 0.7272 (1, 8) | 0.7272 (1, 8) | 0.7532 (3, 2) | 0.7662 (1, 1) | 0.7532 (3, 2) | 0.7532 (3, 2) | 0.7532 (3, 2) | 0.7532 (1, 2) | 0.7532 (3, 2) |
| WDBC | 0.6491 (1, 8) | 0.8947 (29, 11) | 0.6315 (1, 12) | 0.6315 (1, 12) | 0.8596 (1, 10) | 0.8771 (1, 9) | 0.9649 (1, 4) | 0.9649 (1, 4) | 0.9649 (1, 4) | 0.9649 (1, 4) | 0.9824 (15, 1) | 0.9824 (1, 1) | 0.9824 (15, 1) |
| Wine | 0.4888 (1, 11) | 0.8888 (7, 8) | 0.3888 (7, 12) | 0.3333 (1, 13) | 0.8888 (1, 8) | 0.5555 (4, 10) | 1(1, 1) | 1(1, 1) | 1(1, 1) | 1(1, 1) | 1(1, 1) | 1(1, 1) | 1(1, 1) |
| Wholesale | 0.6818 (1, 9) | 0.7954 (5, 8) | 0.6818 (1, 9) | 0.6818 (1, 9) | 0.6818 (1, 9) | 0.6818 (1, 9) | 1(3, 1) | 0.9772 (1, 3) | 1(3, 1) | 0.9772 (1, 3) | 0.9772 (1, 3) | 0.9772 (1, 3) | 0.9772 (1, 3) |
| Average rank | 7.83 | 8.41 | 10.25 | 11.75 | 6.25 | 7.25 | 2.66 | 4.66 | 2.66 | 3.5 | 2.25 | 3.41 | 2.41 |

Totally, out of the 12 experiments, the proposed approaches have earned the first rank in 10 experiments. As you can see, the proposed methods have generally a better average ranking in comparison with the base approaches. Having these observations, we can announce LLE as the best approach in finding the neighborhoods on the manifold in terms of accuracy and using SVM as the classifier.

### 4.4. Representation of the data after reduction

For better visualization of the achievements of the approach after feature mapping and reduction, the data distribution after using the proposed approach is plotted in a 2D space and compared with the original distribution after feature selection and data distribution after DLSR. The plots are shown in Fig. 5.

| Data name | Original 2D view | DLSR | PCA + DLSR |
|---|---|---|---|
| Bupa | 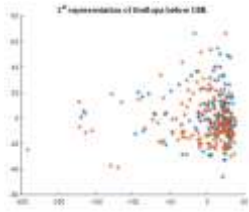 | 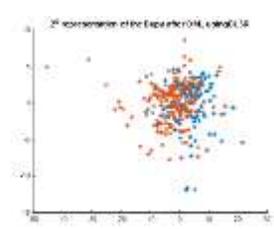 | 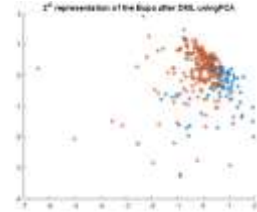 |
| Ionosphere | 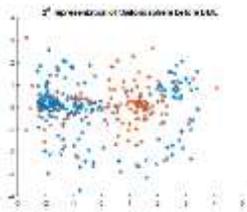 | 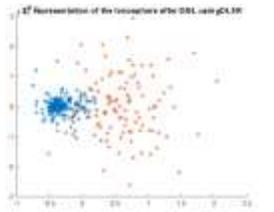 | 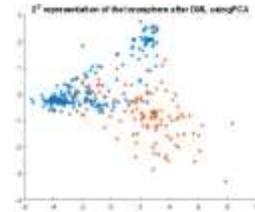 |
| Iris | 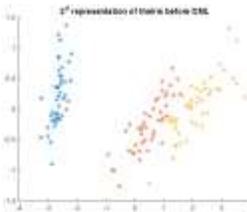 | 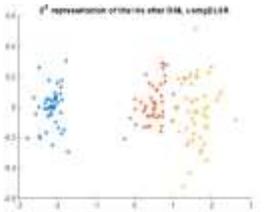 | 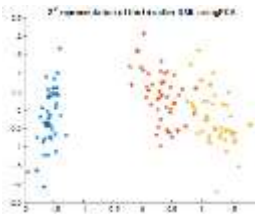 |
| Monks | 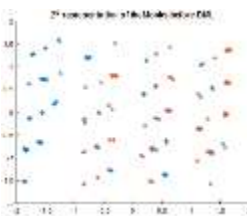 | 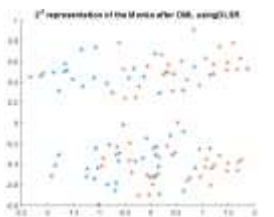 | 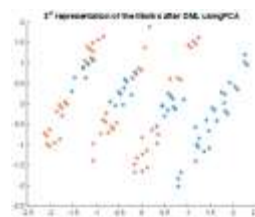 |
| New-thyroid | 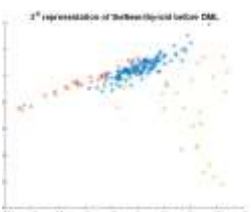 | 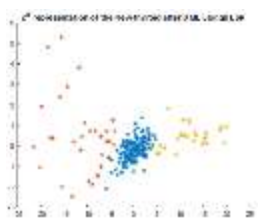 | 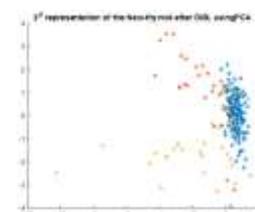 |
| WDBC | 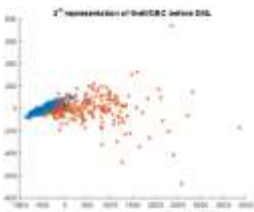 | 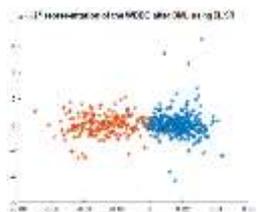 | 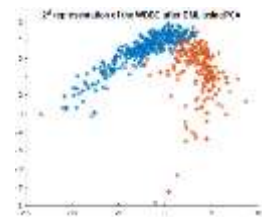 |

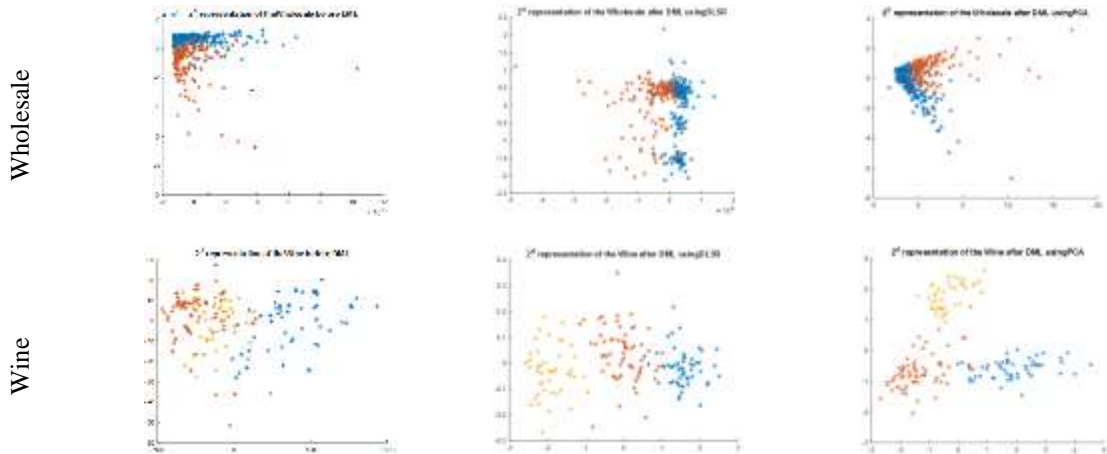

Figure 5. Data distribution visualization after the reduction to a new 2D space using different approaches.

In the above illustrations, PCA is applied as the manifold learning approach, so that the discriminative nature of the approach does not affect the final distribution of samples. As you can see in the illustrations above, the proposed method discriminates the data from different classes far better than the traditional DLSR approach.

### 4.5. Evaluations on the KDD data

In this section we will specifically compare the average results with respect to the confusion matrix of the proposed algorithm and DSLR [25] on KDD dataset which has the highest imbalance ratio among the other datasets studied in this research. The results are shown in Tables 6 to 13.

Note that for the sake of computational facility, as the number of samples in KDD dataset is too great, we have conducted an under sampling before executing the process, i.e. we have only sampled one hundredth of each in the KDD dataset, except the U2R class which is the minority class.

In the following tables, Table 6 denotes DLSR approach results while the others denote the results of the proposed framework using different methods in the manifold learning phase. As seen in Tables 7 to 13, the integer average values for the number of samples correctly classified to each of the classes testifies the predictability and class-wisely equal performance of the proposed methods which signifies the robustness of the approach independent from the fold on which it is tested. Whereas, under the same conditions DLSR method, shown in Table 6, has the non-integer average values for the average number of samples classified to each class in its confusion matrix. This observation on the KDDCup dataset which suffers from high imbalance ratio is the main achievement of the proposed framework on this dataset. However, as can be concluded from these experiments the recall rate of the proposed approach is higher than the DSLR method especially on minority classes (i.e. R2L and U2R).

Table 6. The average confusion matrix of the 10-fold cross validation using DLSR approach on KDD dataset using 7-NN classifier for the best latent dimensionality (i.e. 28).

| Class name | DOS | Normal | Probe | R2L | U2R | Accuracy |
|---|---|---|---|---|---|---|
| DOS | 391.1 | 0.6 | 0 | 0.2 | 0.1 | 0.997704 |
| Normal | 0.2 | 96.5 | 0 | 0.3 | 0 | 0.994845 |
| Probe | 0.1 | 1.2 | 2.7 | 0 | 0 | 0.675 |
| R2L | 0.1 | 0.4 | 0 | 0.4 | 0.1 | 0.4 |
| U2R | 0.1 | 1.1 | 0 | 0.4 | 3.4 | 0.68 |

Table 7. The average confusion matrix of the 10-fold cross validation using the proposed PCA+DSLR dimension reduction on KDD dataset using 7-NN classifier for the best latent dimensionality (i.e. 10).

| Class name | DOS | Normal | Probe | R2L | U2R | Accuracy |
|---|---|---|---|---|---|---|
| DOS | 392 | 0 | 0 | 0 | 0 | 1 |
| Normal | 1 | 95 | 0 | 1 | 0 | 0.979381 |
| Probe | 0 | 0 | 4 | 0 | 0 | 1 |
| R2L | 0 | 0 | 0 | 1 | 0 | 1 |
| U2R | 0 | 1 | 0 | 0 | 4 | 0.8 |

Table 8. The average confusion matrix of the 10-fold cross validation using the proposed LDA+DLSR dimension reduction on KDD dataset using 7-NN classifier for the best latent dimensionality (i.e. 28).

| Class name | DOS | Normal | Probe | R2L | U2R | Accuracy |
|---|---|---|---|---|---|---|
| DOS | 392 | 0 | 0 | 0 | 0 | 1 |
| Normal | 1 | 95 | 0 | 1 | 0 | 0.979381 |
| Probe | 0 | 0 | 4 | 0 | 0 | 1 |
| R2L | 0 | 0 | 0 | 1 | 0 | 1 |
| U2R | 0 | 1 | 0 | 0 | 4 | 0.8 |

Table 9. The average confusion matrix of the 10-fold cross validation using the proposed MDS+DLSR dimension reduction on KDD dataset using 7-NN classifier for the best latent dimensionality (i.e. 1).

| Class name | DOS | Normal | Probe | R2L | U2R | Accuracy |
|---|---|---|---|---|---|---|
| DOS | 392 | 0 | 0 | 0 | 0 | 1 |
| Normal | 1 | 95 | 0 | 1 | 0 | 0.979381 |
| Probe | 0 | 0 | 4 | 0 | 0 | 1 |
| R2L | 0 | 0 | 0 | 1 | 0 | 1 |
| U2R | 0 | 1 | 0 | 0 | 4 | 0.8 |

Table 10. The average confusion matrix of the 10-fold cross validation using the proposed Isomap+DLSR dimension reduction on KDD dataset using 7-NN classifier for the best latent dimensionality (i.e. 1).

| Class name | DOS | Normal | Probe | R2L | U2R | Accuracy |
|---|---|---|---|---|---|---|
| DOS | 392 | 0 | 0 | 0 | 0 | 1 |
| Normal | 1 | 95 | 0 | 1 | 0 | 0.979381 |
| Probe | 0 | 0 | 4 | 0 | 0 | 1 |
| R2L | 0 | 0 | 0 | 1 | 0 | 1 |
| U2R | 0 | 1 | 0 | 0 | 4 | 0.8 |

Table 11. The average confusion matrix of the 10-fold cross validation using the proposed LLE+DLSR dimension reduction on KDD dataset using 7-NN classifier for the best latent dimensionality (i.e. 1).

| Class name | DOS | Normal | Probe | R2L | U2R | Accuracy |
|---|---|---|---|---|---|---|
| DOS | 392 | 0 | 0 | 0 | 0 | 1 |
| Normal | 1 | 95 | 0 | 1 | 0 | 0.979381 |
| Probe | 0 | 0 | 4 | 0 | 0 | 1 |
| R2L | 0 | 0 | 0 | 1 | 0 | 1 |
| U2R | 0 | 1 | 0 | 0 | 4 | 0.8 |

Table 12. The average confusion matrix of the 10-fold cross validation using the proposed KPCA+DLSR dimension reduction on KDD dataset using 7-NN classifier for the best latent dimensionality (i.e. 1).

| Class name | DOS | Normal | Probe | R2L | U2R | Accuracy |
|---|---|---|---|---|---|---|
| DOS | 392 | 0 | 0 | 0 | 0 | 1 |
| Normal | 1 | 95 | 0 | 1 | 0 | 0.979381 |
| Probe | 0 | 0 | 4 | 0 | 0 | 1 |
| R2L | 0 | 0 | 0 | 1 | 0 | 1 |
| U2R | 0 | 1 | 0 | 0 | 4 | 0.8 |

Table 13. The average confusion matrix of the 10-fold cross validation using the proposed Autoencoder+DLSR approach on KDD dataset using 7-NN classifier for the best latent dimensionality (i.e. 1).

| Class name | DOS | Normal | Probe | R2L | U2R | Accuracy |
|---|---|---|---|---|---|---|
| DOS | 392 | 0 | 0 | 0 | 0 | 1 |
| Normal | 1 | 95 | 0 | 1 | 0 | 0.979381 |
| Probe | 0 | 0 | 4 | 0 | 0 | 1 |
| R2L | 0 | 0 | 0 | 1 | 0 | 1 |
| U2R | 0 | 1 | 0 | 0 | 4 | 0.8 |

Also, Table 14 shows the best results of the proposed approach which is achieved using SVM classifier and Autoencoder as the manifold learning approach. This experiment is also performed using 10 fold cross validation. The same results as above are observed in the following table while the classification accuracy of the approach in different classes (even the minority classes) is considerably high.

Table 14. The average confusion matrix of the 10-fold cross validation using the proposed Autoencoder+DLSR approach on KDD dataset using the SVM classifier for the best latent dimensionality (i.e. 9).

| Class name | DOS | Normal | Probe | R2L | U2R | Accuracy |
|---|---|---|---|---|---|---|
| DOS | 392 | 0 | 0 | 0 | 0 | 1 |
| Normal | 0 | 97 | 0 | 0 | 0 | 1 |
| Probe | 0 | 0 | 4 | 0 | 0 | 1 |
| R2L | 0 | 0 | 0 | 1 | 0 | 1 |
| U2R | 0 | 1 | 0 | 0 | 4 | 0.8 |

To have a better representation of the achievements of the proposed approach on KDD dataset compared with some recent and the state of the art approaches, we also had a class-wise comparison between the best accuracy results of the proposed method and the TVCPSO [27] and CANN [28] approaches on the KDD dataset. The TVCSPO [27] approach is designed to propose a framework for the intrusion detection using an adaptive, robust with precise optimization novel approach called the Time-varying chaos particle swarm optimization which is used for concurrent parameter setting and feature selection for the multiple criteria linear programming (MCLP) and the SVM classification. In this approach a weighted objective function is used which handles a tradeoff between the detection rate maximization and false alarm rate minimization, by considering the number of features. Furthermore, in this approach, in order to make the particle swarm optimization faster in finding the global optimal point and avoid the local optima, the chaos is concept is adopted in the PSO and the time varying inertia weight and the time varying acceleration coefficient is introduced.

CANN [28], proposes an approach called the cluster center and nearest neighbor. In this approach, two distances are measured and aggregated, the first one is based on the distance between each data sample and its cluster center, and the second one is based on the distance between each data point and its nearest neighbor form the same cluster. This new and one dimensional representation of data points is used for the intrusion detection by a KNN classifier.

Table 15 shows the results of the mentioned approaches in comparison with the proposed approach. As seen in Table 15, the proposed method as is specified in Table 14, performs considerably better in terms of accuracy in comparison with the recently proposed methods [27], [28] on the KDD dataset. Other than an improvement on majority classes such as DOS and Normal, the proposed approach is highly efficient in identifying the minority classes such as R2L and U2R which is the main drawback of previous approaches on this datasets.

Table 15. A comparison between the accuracy of the proposed method and some other recent works on different classes of the KDD based on the 10-fold cross validation.

| Class name | Proposed Method using AE and SVM | CANN [28] | TVCPSO-MCLP [27] | TVCPSO-SVM [27] |
|---|---|---|---|---|
| DOS | **1** | 0.9968 | 0.9864 | 0.9884 |
| Normal | **1** | 0.9704 | 0.9759 | 0.9913 |
| Probe | **1** | 0.8761 | 0.8790 | 0.8929 |
| R2L | **1** | 0.5702 | 0.7508 | 0.6784 |
| U2R | **0.8** | 0.385 | 0.5962 | 0.4038 |
| Average | **0.96** | 0.7597 | 0.83766 | 0.79096 |

## 5. Conclusions and future works

In this research a novel method for distance metric learning with the aim of preserving the local neighborhoods between similar data points and also covering data imbalance problem has been proposed and the implementation steps and its experimental results in comparison with other distance metric learning and dimensionality reduction algorithms has been evaluated. In the proposed method, it has been tried to first learn the neighborhoods between the data points based on their neighborhood relations on the manifold. For each data point, two neighborhoods with same number of members consisting of the similar and dissimilar data points to the given point are created. Consequently, distance metric learning is performed with the goal of making the similar points nearer to the given data point and to push back the dissimilar data away from it. Finally, thanks to learned transformation matrix, data are mapped to the similarity space and then the classification is preformed using k-NN and SVM classifiers. The evaluations are performed on 12 datasets with different sizes and imbalance ratio specially the KDD, which resulted in significant results based on the three criteria of accuracy, sensitivity and specificity.

In future we would like to have a study on different approaches of data sampling specially the graph based prototype selection approaches which preserve the local structures of the data. Besides, as for the graph prototype selection, we need to calculate the appropriate distance between different graphs in order to select the most expressive ones. Therefore, another area that we could invest on in the future is to study on the effect of using different graph editing distances on the graph based prototype selection.

An analysis on the selection of the most appropriate manifold learning approaches (as different manifold learning approaches result in differently manifolds) could have an extensive impact on the improvement of the learned distance metric. To do this, we would like to have a study on the effect of using the deep neural networks e.g., convolutional neural networks and generative adversarial networks as the learning approaches and analyze their results in comparison with the existing manifold learning approaches.